%% file: WACV_IEGAN/egpaper_for_review.tex
\documentclass[10pt,twocolumn,letterpaper]{article}

\usepackage{wacv}
\usepackage{times}
\usepackage{epsfig}
\usepackage{graphicx}
\usepackage{amsmath}
\usepackage{amssymb}
\usepackage{float}
\usepackage{graphicx}
\usepackage{amsmath} 
\usepackage{amssymb}
\usepackage{color}
\usepackage[table,xcdraw]{xcolor}
\usepackage{subfigure}

\usepackage{mathtools}
\usepackage{multirow}
\usepackage{amssymb}
\usepackage[T1]{fontenc}
\usepackage{babel}
\usepackage{breqn}

\graphicspath{{images/}}

\input{macro.tex}

\wacvfinalcopy 

\def\wacvPaperID{623} 

\ifwacvfinal\fi
\begin{document}

\title{IEGAN: Multi-purpose Perceptual Quality Image Enhancement Using Generative Adversarial Network}

\author{
\vspace{0.5em}
  Soumya Shubhra Ghosh\textsuperscript{1}, Yang Hua\textsuperscript{1}, 
  Sankha Subhra Mukherjee\textsuperscript{2}, Neil Robertson\textsuperscript{1,2}\\
  \vspace{0.5em}
  \textsuperscript{1}EEECS/ECIT, Queen's University Belfast \hspace{2em} \textsuperscript{2}Anyvision\\
  \texttt{\{sghosh02,y.hua,n.robertson\}@qub.ac.uk, rick@anyvision.co}
}

\maketitle
\ifwacvfinal\thispagestyle{empty}\fi

\begin{abstract}
Despite the breakthroughs in quality of image enhancement, an end-to-end solution for simultaneous recovery of the finer texture details and sharpness for degraded images with low resolution is still unsolved. Some existing approaches focus on minimizing the pixel-wise reconstruction error which results in a high peak signal-to-noise ratio. The enhanced images fail to provide high-frequency details and are perceptually unsatisfying, i.e., they fail to match the quality expected in a photo-realistic image. In this paper, we present Image Enhancement Generative Adversarial Network (IEGAN), a versatile framework capable of inferring photo-realistic natural images for both artifact removal and super-resolution simultaneously. Moreover, we propose a new loss function consisting of a combination of reconstruction loss, feature loss and an edge loss counterpart. The feature loss helps to push the output image to the natural image manifold and the edge loss preserves the sharpness of the output image. The reconstruction loss provides low-level semantic information to the generator regarding the quality of the generated images compared to the original. Our approach has been experimentally proven to recover photo-realistic textures from heavily compressed low-resolution images on public benchmarks and our proposed high-resolution World100 dataset.
\end{abstract}

\section{Introduction}
\label{sec:intro}
%
Photo-Realistic image enhancement is challenging but highly demanded in real-world applications. Image enhancement can be broadly classified into two domains: super-resolution (SR) and artifact removal (AR). The task of estimating a high-resolution image from its low-resolution (LR) counterpart is the SR and estimating an artifact-free sharp image from its corrupted counterpart is the AR. 

\begin{figure}[t]
\centering
\begin{minipage}{.15\textwidth}
  \centering
  \includegraphics[width=0.88\linewidth]{Untitled.jpg}
  \\Input
  \figfigvspace
\end{minipage}%
\begin{minipage}{.15\textwidth}
  \centering
  \includegraphics[width=0.88\linewidth]{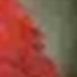}
  Bicubic
  \figfigvspace
\end{minipage}
\begin{minipage}{.15\textwidth}
  \centering
  \includegraphics[width=0.88\linewidth]{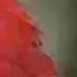}
  \small{ARCNN+SRGAN}
  \figfigvspace
\end{minipage}%

\begin{minipage}{.15\textwidth}
  \centering
  \includegraphics[width=0.88\linewidth]{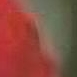}
  \small{SRGAN+ARCNN}
  \figfigvspace
\end{minipage}
\begin{minipage}{.15\textwidth}
  \centering
  \includegraphics[width=0.88\linewidth]{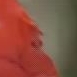}
  \small{IEGAN}
  \figfigvspace
\end{minipage}%
\begin{minipage}{.15\textwidth}
  \centering
  \includegraphics[width=0.88\linewidth]{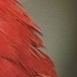}
  Ground Truth
  \figfigvspace
\end{minipage}
\figfigvspace
\caption{End-to-end AR+SR of color images. The input image is degraded to 10\% of its original quality and reduced by a factor of 4. The output image from proposed IEGAN shows better reconstruction and sharper edges compared to the other algorithms. Best viewed in pdf.}
\label{fig:ARSRmotivation}
\end{figure}
%
The AR problem is particularly prominent for highly compressed images and videos, for which texture detail in the reconstructed images is typically absent. The same problem persists for SR as well. One major problem with the current state-of-the-art is that there does not exist any end-to-end network which can solve the problem of AR and SR simultaneously, thus requiring two different algorithms to be applied on the image if both AR and SR are desirable. This is the most common problem in images on the Internet (for instance in Twitter, Instagram etc.) or for object recognition and classification from the surveillance videos where some people/objects are typically far away from the camera and appear small in the images. A simultaneous super-resolution and artifact-removal is highly useful in these scenarios and we have explored this possibility in this paper. To cope with the problem of generating high perceptual quality images, different approaches have been proposed \cite{feifeilistyletransfer,perceptualsimilarity,bruna2015super}. These approaches deal either with SR or with AR, but not both.

 Supervised image enhancement algorithms \cite{SRCNN,ARCNN,L4L8} generally tries to minimize the mean squared error (MSE) between the target high-resolution (HR) image and the ground truth, thus maximizing the PSNR. However, the ability of MSE to capture perceptually relevant differences, such as high texture detail, is insufficient as they are defined based on pixel-wise image differences. This leads to an image having an inferior perceptual quality. Recently, deep learning has shown impressive results. In particular, the Super Resolution Convolutional Neural Network (SRCNN) proposed by Dong et al. \cite{SRCNN} shows the potential of an end-to-end deep convolutional network in SR. Ledig {\it et al.} \cite{SRGAN} presented a framework called SRGAN which is capable of generating photo-realistic images for $4\times$ up-scaling factors, but there are several problems of this framework when used for SR in conjunction with an AR framework. Dong {\it et al.} \cite{ARCNN} discovered that SRCNN directly applied for compression artifact reduction leads to undesirable noisy patterns, thus proposing a new improved model called Artifacts Reduction Convolutional Neural Networks (ARCNN), which showed better performance. Svoboda {\it et al.} \cite{L4L8} proposed the L4 and L8 architecture which has better results compared to ARCNN but still failed to completely remove all the artifacts for highly compressed JPEG image. A major drawback for all the successful methods till date is that all the proposed methods work on the Luma channel (channel Y in YCbCr color space which is monochrome), but none of them reports the performance on color images, although AR in color images is more relevant. As per our knowledge, till date, a versatile robust algorithm which solves all kind of image enhancement problems is yet to be proposed. 

In this paper we propose a novel Image Enhancing Generative Adversarial Network (IEGAN) using U-net like generator with skip connections and an autoencoder-like discriminator. This is a multi-purpose image enhancement network which is capable of removing artifacts and super-resolving with high sharpness and details in an end-to-end manner, simultaneously, within a single network. Our main contributions are summarized as follows:

\begin{itemize}
\item[$\bullet$] We propose the first end-to-end network called Image Enhancement Generative Adversarial Network (\textbf{IEGAN}) which can solve the problem of SR and AR simultaneously. Our proposed network is able to generate photo-realistic images from low-resolution images corrupted with artifacts, i.e., it acts as a unified framework which simultaneously super-resolves the image and recovers it from the compression artifacts. 

\item[$\bullet$] We propose a new and improved perceptual loss function which is the sum of the reconstruction loss of the discriminator, the feature loss from the VGG network \cite{vgg} and the edge loss from the edge detector. This novel loss function preserves the sharpness of the enhanced image which is often lost during enhancement.

\item[$\bullet$] We also create a benchmark dataset named \textbf{World100} for testing the performance of our algorithms on high-resolution images. 
\end{itemize}



\section{Related Work}

\subsection{Image Artifact Removal}

AR of compressed images has been extensively dealt with in the past. In the spatial domain, different kinds of filters \cite{deblocking80s,adaptivedeblockingfilter,nonlocaladaptivedeblocking} have been proposed to adaptively deal with blocking artifacts in specific regions. In the frequency domain, wavelet transform has been utilized to derive thresholds at different wavelet scales for deblocking and denoising \cite{deblockingwavelet,dctwaveletdeblocking}. However, the problems with these methods are that they could not reproduce sharp edges, and tend to have overly smooth texture regions. In the recent past, JPEG compression AR algorithms involving deep learning has been proposed. Designing a deep model for AR requires a deep understanding of the different artifacts. Dong {\it et al.} \cite{ARCNN} showed that directly applying the SRCNN architecture for JPEG AR resulted in undesired noisy patterns in the reconstructed image, and thus proposed a new improved model. Svoboda {\it et al.} \cite{L4L8} proposed a novel method of image restoration using convolutional networks that had a significant quality advancement compared to the then state-of-the-art methods. They trained a network with eight layers in a single step and in a relatively short time by combining residual learning, skip architecture, and symmetric weight initialization.

\subsection{Image Super-resolution}

Initially filtering approaches were used for SR. They are usually very fast but with overly smooth textures, thus losing a lot of details. Methods focusing on edge-preservation \cite{edgeSR1,edgeSR2} also fail to produce photo-realistic images. Recently convolutional neural network (CNN) based SR algorithms have shown excellent performance. Wang {\it et al.} \cite{wang2015deep} showed that sparse coding model for SR can be represented as a neural network and improved results can be achieved. Dong {\it et al.} \cite{SRCNN} used bicubic interpolation to upscale an input image and trained a three-layer deep fully convolutional network end-to-end achieving state-of-the-art SR performance. Dong {\it et al.} \cite{acceleratingSR} and Shi {\it et al.} \cite{pixelshuffle} demonstrated that upscaling filters can be learnt for SR for increased performance. The studies of Johnson {\it et al.} \cite{feifeilistyletransfer} and Bruna {\it et al.} \cite{bruna2015super} relied on loss functions which focus on perceptual similarity to recover HR images which are more photo-realistic. A recent work by Ledig {\it et al.} \cite{SRGAN} presented the first framework capable of generating photo-realistic images for $4\times$ upscaling factors. Sajjadi {\it et al.} proposed a novel application of automated texture synthesis in combination with a perceptual loss which focuses on creating realistic textures.

\subsection{Loss Functions}

Pixel-wise loss functions like MSE or L1 loss are unable to recover the lost high-frequency details in an image. These loss functions encourage finding pixel-wise averages of possible solutions, which are generally smooth but have poor perceptual quality \cite{bruna2015super,perceptualsimilarity,feifeilistyletransfer,SRGAN}. Ledig {\it et al.} \cite{SRGAN} illustrated the problem of minimizing MSE where multiple plausible solutions with high texture details are averaged creating a smooth reconstruction. Johnson {\it et al.} \cite{feifeilistyletransfer} and Bruna {\it et al.} \cite{bruna2015super} proposed extracting the features from a pre-trained VGG network instead of using pixel-wise error. They proposed a perceptual loss function based on the Euclidean distance between feature maps extracted from the VGG19 \cite{vgg} network. Ledig {\it et al.} \cite{SRGAN} proposed a GAN-based network optimized for perceptual loss which are more invariant to changes in pixel space, obtaining better visual results.

\subsection{Perceptual Image Quality Evaluation}

Evaluating the perceptual quality of an image is tricky because most of the statistical measures does not well reflect the human perception. Ledig {\it et al.} \cite{SRGAN} has shown this in their work that images with high PSNR does not necessarily mean a perceptually better image. Same applies to Structural Similarity (SSIM) as well. Xue {\it et al.} \cite{GMSD} presented an effective and efficient image quality assessment model called Gradient Magnitude Similarity Deviation (GMSD) which they claimed to have favorable performance in terms of both perceptual quality and efficiency. A statistical analysis on image quality measures conducted by Kundu {et al.} reported that GMSD \cite{ImageQMetrics}  showed a high correlation with human visual system. A very recent work by Reisenhofer {\it et al.}  presents a similarity measure for images called Haar wavelet-based Perceptual Similarity Index (HaarPSI) \cite{HaarPSI} that aims to correctly assess the perceptual similarity between two images with respect to a human viewer. It achieves higher correlations with human opinion scores on large benchmark databases in almost every case and is probably the best perceptual similarity metric available in the literature.

Taking these into account, the similarity metrics we have selected for evaluating the performance are GMSD \cite{GMSD} and HaarPSI \cite{HaarPSI}. We have also calculated the PSNR and SSIM \cite{SSIM} for a fair comparison with other algorithms.

\section{Our Approach}
\label{sec:method}

In this paper, we aim to estimate a sharp and artifact free image $I^{HR}$ from an image $I^{LR}$ which is either low-resolution or corrupted with artifacts or both. Here $I^{HR}$ is the enhanced version of its degraded counterpart $I^{LR}$. For an image with $C$ channels, we describe $I^{LR}$ by a real-valued tensor of size $W\times H\times C$ and $I^{HR}$ and $I^{GT}$ by $\rho W\times \rho H\times \rho C$ respectively, where $I^{GT}$ is the ground truth image and $\rho=2^p$ where $p\in\{0,1,2,...\}$.

In order to estimate the enhanced image for a given low-quality image, we train a generator network as a feed-forward CNN $G_{\theta_{G}}$ parametrized by $\theta_G$. Here $\theta_G = {W_1:L; b_1:L}$ denotes the weights and biases of a L-layer deep network and is obtained by optimizing a loss function $F_{loss}$. The training is done using two sets of $n$ images $\{I^{GT}_i:i=1,2,...,n\}$ and $\{I^{LR}_j:j=1,2,...,n\}$ such that $I^{GT}_i = G_{\theta_{G}}(I^{LR}_j)$ (where $I^{GT}_i$ and $I^{LR}_j$ are corresponding pairs) and by solving 

\begin{equation}
\label{eq:optimize_loss}
{\hat{\theta}}_G = \operatorname*{arg\,min}_{\theta_G} \dfrac{1}{N}\sum_{\substack{i,j=1\\i=j}}^{n}F_{loss}(I^{GT}_i, G_{\theta_{G}}(I^{LR}_j))
\end{equation}

Following the work of Goodfellow {\it et al.} \cite{GAN} and Isola {\it et al.} \cite{pix2pix}, we also add a discriminator network $D_{\theta_{D}}$ to assess the quality of images generated by the generator network $G_{\theta_{G}}$. 

The generative network is trained to generate the target images such that the difference between the generated images and the ground truth are minimized. While training the generator, the discriminator is trained in an alternating manner such that the probability of error of the discriminator (between the ground truth and the generated images) is minimized. With this adversarial min-max game, the generator can learn to create solutions that are highly similar to real images. This also encourages perceptually superior solutions residing in the manifold of natural images.

\subsection{Network Architecture}
\label{subsec:architecture}

\begin{figure*}[t]
\begin{center}
\begin{minipage}{\textwidth}
  \centering
  \includegraphics[width=\linewidth]{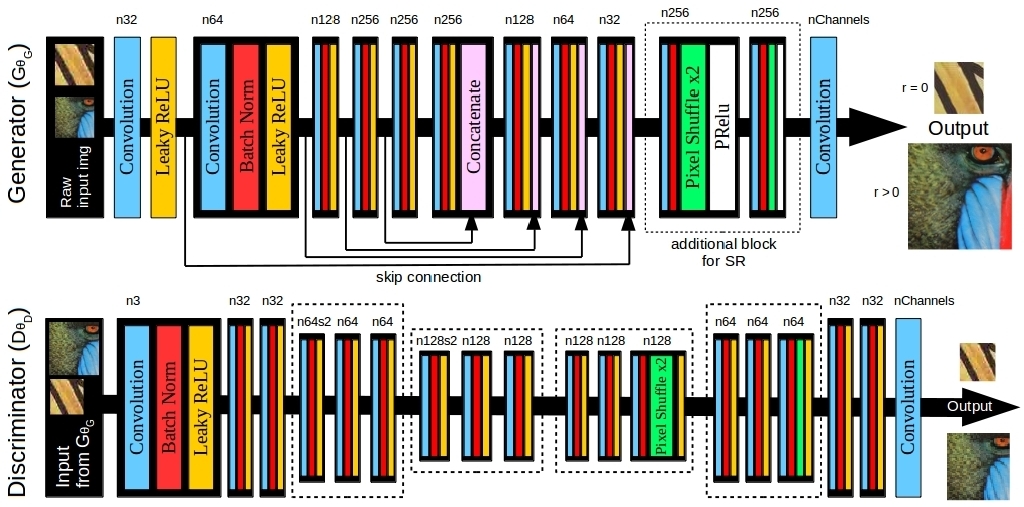}
\end{minipage}
\end{center}
\caption{The overall architecture of our proposed network. The convolution layers of the generator have a kernel size $3\times3$ and stride is 1. Number of filters for each layer is indicated in the illustration, e.g., n32 refers to 32 filters. For the discriminator network, the stride is 1 except for the layers which indicates that the stride is 2, e.g., n64s2 refers to 64 filters and stride=2}
\label{fig:architecture}
\end{figure*}



We follow the architectural guidelines of GAN proposed by Radford {\it et al.} \cite{DCGAN}. For the generator we use convolutional layers with small $3\times3$ kernels and stride=1 followed by batch-normalization layers \cite{InceptionV2batchnorm} and Leaky ReLU \cite{leakyrelu} as the activation function. The number of filters per convolution layer is indicated in Figure \ref{fig:architecture}. 

For image enhancement problems, even though the input and output differ in appearance, both are actually renderings of the same underlying structure. Therefore, the input is more or less aligned with the output. We design the generator architecture keeping these in mind. For many image translation problems, there is a lot of low-level information shared between the input and output, and it will be helpful to pass this information directly across the network. Ledig {\it et al.} had used residual blocks and a skip connection in their SRGAN \cite{SRGAN} framework to help the generator carry this information. However, we found that it is more useful to add skip connections following the general shape of a U-Net \cite{unet}. Specifically, we add skip connections between each layer $n$ and layer $L-n$, where $L$ is the total number of layers. Each skip connection simply concatenates all channels at layer $n$ with those at layer $L-n$. The proposed deep generator network $G_{\theta_{G}}$ is illustrated in Figure \ref{fig:architecture}. The generator has an additional block containing two sub-pixel convolution layers immediately before the last layer for cases where $p>0$, i.e., where the size of the output is greater than the input. These layers are called pixel-shuffle layers, as proposed by Shi {\it et al.} \cite{pixelshuffle}. Each pixel shuffle layer increases the resolution of the image by $2\times$. In Figure \ref{fig:architecture}, we show two such layers which super-resolves the image by $4\times$. If $p=0$, we do not need any such block since the size of the output image is equal to the input image.

The discriminator in our framework is very crucial for the performance and is designed in the form of an autoencoder. Thus the output of the autoencoder is the reconstructed image of its input which is the ground truth or the generator output. This helps the discriminator to pass back a lot of semantic information to the generator regarding the quality of the generated images, which is not possible with a binary discriminator. Our proposed discriminator contains eighteen convolutional layers with an increasing number of $3\times3$ filter kernels. The specific number of filters are indicated in Figure \ref{fig:architecture}. Strided convolutions with stride=2 are used to reduce the feature map size, and pixel-shuffle layers \cite{pixelshuffle} are used to increase them. The overall architecture of the proposed framework is shown in Figure \ref{fig:architecture} in details.

\subsection{Loss Function}
\label{subsec:lossfunction}

The performance of our network highly varies with different loss functions. Thus a proper loss function is critical for the performance of our generator network. We improve on Johnson {\it et al.} \cite{feifeilistyletransfer}, Bruna {\it et al.} \cite{bruna2015super} and Ledig {\it et al.} \cite{SRGAN} by adding an edge loss counterpart and the discriminator reconstruction loss to design a loss function that can asses an image with respect to perceptual features instead of minimizing pixel-wise difference. The absence of the edge loss and the reconstruction loss counterpart in SRGAN is an important reason why it fails to produce sharp images during AR+SR. Adding these helps to produce sharp output images even after removal of artifacts and $4\times$ up-scaling.

\subsubsection{Feature Loss}
\label{subsubsec:featureloss}

We choose the feature loss based on the ReLU activation layers of the pre-trained 19 layer VGG network described in Simonyan and Zisserman \cite{vgg}. This loss is described as VGG loss by Ledig {\it et al.} \cite{SRGAN} and is mathematically expressed as

\begin{dmath}
\label{eq:vggloss}
C_{loss}^{VGG_{i.j}}(I^{GT}, G_{\theta_{G}}(I^{LR})) = \dfrac{1}{W_{i,j}H_{i,j}}\sum_{x=1}^{W_{i,j}}\sum_{y=1}^{H_{i,j}}(\phi_{i,j}(I^{GT})_{x,y} - \phi_{i,j}(G_{\theta_{G}}(I^{LR}))_{x,y})^2
\end{dmath}

\noindent where $\phi_{i,j}$ is the feature map obtained by the $j^{th}$ convolution (after activation) before the $i^{th}$ max-pooling layer within the pre-trained VGG19 network, and W and H represents the width and height of input image, respectively. 

\subsubsection{Edge Loss}
\label{subsubsec:edgeloss}

Preservation of edge information is very important for the generation of sharp and clear images. Thus we add an edge loss to the feature loss counterpart. There are several edge detectors available in the literature, and we have chosen to design our edge loss function using the state of the art edge detection algorithm proposed by Xie and Tu called Holistically-nested Edge Detection (HED) \cite{hed} and the classical Canny edge detection algorithm \cite{CannyEdge} due to its effectiveness and simplicity. Experimental results prove that the Canny algorithm provides similar results for the preservation of sharpness but with greater speed and fewer resource requirements compared to HED. The detailed comparison results have been further discussed in Section \ref{subsec:investigation}. For the Canny algorithm, a Gaussian filter of size $3\times3$ with $\sigma=0.3$ was chosen as the kernel. This loss is mathematically expressed as
\begin{dmath}
\label{eq:edgeloss}
E_{loss}^{edge}(I^{GT}, G_{\theta_{G}}(I^{LR})) = \dfrac{1}{WH}\sum_{x=1}^{W}\sum_{y=1}^{H} \bigg| \Theta(I^{GT})_{x,y} - \Theta(G_{\theta_{G}}(I^{LR}))_{x,y} \bigg|
\end{dmath}
where $\Theta$ is the edge detection function.

\subsubsection{Reconstruction Loss}
\label{discriminatortraining}

Unlike most other algorithms, our discriminator provides a reconstructed image of the discriminator input. Modifying the idea of Berthelot {\it et al.} \cite{began}, we design the discriminator to differentiate between the loss distribution of the reconstructed real image and the reconstructed fake image. Thus we have the reconstruction loss function as
\begin{dmath} 
\label{eq:obj_fn}
\mathcal{L}_D = |\mathcal{L}_D^{real} - k_t\times\mathcal{L}_D^{fake}| 
\end{dmath}
where $\mathcal{L}_D^{real}$ is the loss distribution between the input ground truth image and the reconstructed output of the ground truth image, mathematically expanded as
\begin{dmath}
\mathcal{L}_D^{real} = r \times E_{loss}^{edge}(D_{\theta_{D}}(I^{GT}), D_{\theta_{D}}(G_{\theta_{G}}(I^{GT}))) + (1 - r) \times C_{loss}^{VGG_{i.j}}(D_{\theta_{D}}(I^{GT}), D_{\theta_{D}}(G_{\theta_{G}}(I^{GT})))
\end{dmath}
$\mathcal{L}_D^{fake}$ is the loss distribution between the generator output image and the reconstructed output of the same, expanded as
\begin{dmath}
\mathcal{L}_D^{fake} = 
r \times E_{loss}^{edge}(D_{\theta_{D}}(I^{LR}), D_{\theta_{D}}(G_{\theta_{G}}(I^{LR}))) + (1 - r) \times C_{loss}^{VGG_{i.j}}(D_{\theta_{D}}(I^{LR}), D_{\theta_{D}}(G_{\theta_{G}}(I^{LR})))
\end{dmath}
and $k_t$ is a balancing parameter at the $t^{th}$ iteration which controls the amount of emphasis put on $\mathcal{L}_D^{fake}$.
\begin{dmath}
k_{t+1} = k_t + \lambda(\gamma\mathcal{L}_D^{real} - \mathcal{L}_D^{fake}) \text{ $\forall$ step } t
\end{dmath}
$\lambda$ is the learning rate of $k$ which is set as $10^{-3}$ in our experiments. Details about this can be found in \cite{began}.

\subsubsection{Final Loss Function}
We formulate the final perceptual loss $F_{loss}$ as the weighted sum of the feature loss $C_{loss}$ and the edge loss $E_{loss}$ component added to the reconstruction loss such that
\begin{equation}
\label{eq:lossfunc}
F_{loss} = r\times E_{loss} + (1-r)\times C_{loss} + \mathcal{L}_D 
\end{equation}

\noindent Substituting the values from Equation \ref{eq:vggloss}, \ref{eq:edgeloss} and \ref{eq:obj_fn}, we have

\begin{dmath}
\label{eq:fulllossfunc}
F_{loss} = r\times \dfrac{1}{WH}\sum_{x=1}^{W}\sum_{y=1}^{H} \bigg| \Theta(I^{GT})_{x,y} - \Theta(G_{\theta_{G}}(I^{LR}))_{x,y} \bigg| + (1-r)\times \dfrac{1}{W_{i,j}H_{i,j}}\sum_{x=1}^{W_{i,j}}\sum_{y=1}^{H_{i,j}}(\phi_{i,j}(I^{GT})_{x,y} - \phi_{i,j}(G_{\theta_{G}}(I^{LR}))_{x,y})^2 + |\mathcal{L}_D^{real} - k_t\times\mathcal{L}_D^{fake}|
\end{dmath}

\noindent The value of $r$ has been decided experimentally.

\section{Experiments}
\subsection{Data, Evaluation Metrics and Implementation Details}
\label{subsec:exp_data}

To validate the performance of AR, we test our framework on the LIVE1 \cite{LIVE1} dataset (29 images) which is the most popular benchmark for AR. For SR, we evaluate the performance using the benchmark datasets Set14 \cite{Set14} (14 images) and BSD100 (100 images) which is a testing set of BSD300 \cite{BSD300}. For simultaneous AR+SR, we conduct the evaluation using the LIVE1\cite{LIVE1} and the World100 dataset. Our proposed World100 dataset contains 100 high-resolution photos representing photographs commonly found. The photographs have all the characteristics e.g., texture, color gradient, sharpness etc., which are required to test any image enhancement algorithm. All the results reported for AR experiments for all the datasets are performed by degrading JPEG images to a quality factor of 10\% (i.e., 90\% degradation), the SR experiments are performed with an upscaling factor of 4, and for AR+SR, the dataset is degraded to quality factor of 10\% and the resolution is reduced by a factor of 4, which corresponds to a 16 times reduction in image pixels. 

We trained all networks on an NVIDIA DGX-1 using a random sample of 60,000 images from the ImageNet dataset \cite{imagenet}. For each mini-batch, we cropped the random $96\times96$ HR sub-images of distinct training images for SR, $256\times256$ for AR, and $128\times128$ for AR+SR. Our generator model can be applied to images of arbitrary size as it is a fully convolutional network. We scaled the range of the image pixel values to [-1, 1]. During feeding the outputs to the VGG network for calculating loss function, we scale it back to [0, 1] since VGG network inherently handles image pixels in the range [0, 1]. For optimization we use Adam \cite{adam} with $\beta _1$ = 0.9. The value of $r$ in Equation \ref{eq:lossfunc} is selected as 0.4. The network was trained with a learning rate of $10^{-4}$ and with $5\times10^4$ update iterations. Our implementation is based on TensorFlow.

\begin{figure}[t!]
\centering
\begin{minipage}{.15\textwidth}
  \centering
  \includegraphics[width=0.95\linewidth]{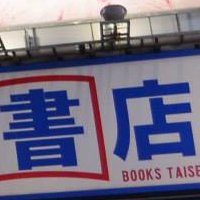}
\end{minipage}%
\begin{minipage}{.15\textwidth}
  \centering
  \includegraphics[width=0.95\linewidth]{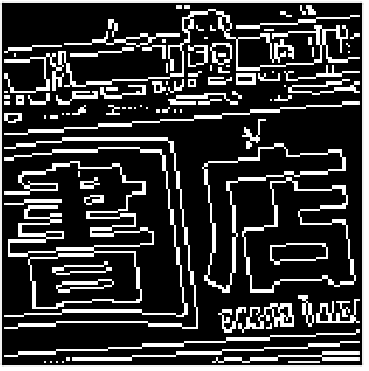}
\end{minipage}
\begin{minipage}{.15\textwidth}
  \centering
  \includegraphics[width=0.95\linewidth]{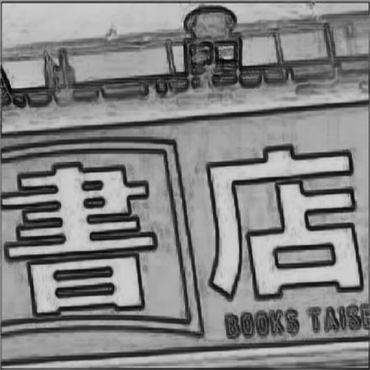}
\end{minipage}
\figfigvspace
\caption{Comparison of edge detection for Canny and HED. Left to Right - Image, edge output of Canny, edge output of HED. Best viewed in pdf.}
\label{fig:edge}
\end{figure}

\setlength{\tabcolsep}{4pt}
\begin{table}[]
\begin{center}
\begin{tabular}{llllll}
\hline
\noalign{\smallskip} 
\multicolumn{5}{c}{\textbf{AR}} \\ 
\hline\noalign{\smallskip}
\rowcolor[HTML]{dfdfdf} 
Loss  & PSNR & SSIM   & GMSD $\downarrow$ & HaarPSI\\
\noalign{\smallskip}
\hline
\noalign{\smallskip}
VGG+Canny     & \textbf{27.31}  & 0.8124 & \textbf{0.0685} & \textbf{0.7533}  \\ 
VGG+HED    & 27.27  & \textbf{0.8180}     & 0.0705   & 0.7481  \\
\noalign{\smallskip}
\hline
\noalign{\smallskip} 
\multicolumn{5}{c}{\textbf{SR-4x}}  \\ 
\hline\noalign{\smallskip}
\rowcolor[HTML]{dfdfdf} 
Loss  & PSNR & SSIM  & GMSD $\downarrow$ & HaarPSI\\ 
\noalign{\smallskip}
\hline
\noalign{\smallskip}
VGG+Canny   & \textbf{25.03}  & 0.7346 & \textbf{0.0850} & \textbf{0.7297} \\
VGG+HED   & 25.03  & \textbf{0.7457}  & 0.0861 & 0.7279\\
\hline
\end{tabular}
\end{center}
\caption{Performance of SR and AR for different edge detectors. AR is evaluated on LIVE1 and SR on Set14. For GMSD, lower value is better.}
\label{table:combiResults}
\end{table}
\setlength{\tabcolsep}{1.4pt}

\setlength{\tabcolsep}{4pt}
\begin{table}[t!]
\begin{center}


\begin{tabular}{lllll}
\hline\noalign{\smallskip}
\multicolumn{5}{c}{\textbf{G + Dv1}}  \\
\hline\noalign{\smallskip}
\rowcolor[HTML]{dfdfdf} 
Loss & PSNR & SSIM & GMSD $\downarrow$ & HaarPSI\\
\noalign{\smallskip}
\hline
\noalign{\smallskip}
VGG           & 27.12           & 0.801          & 0.074 & 0.737 \\ 
L1            & 27.45           & 0.803          & 0.079 & 0.725 \\ 
Canny+VGG     & 27.31           & 0.803  & \textbf{0.073} & \textbf{0.739} \\
Canny+L1 & \textbf{27.74} & \textbf{0.811}       & 0.075 & 0.738  \\
\hline\noalign{\smallskip}
\multicolumn{5}{c}{\textbf{G + Dv2}} \\
\hline\noalign{\smallskip}
\rowcolor[HTML]{dfdfdf} 
Loss & PSNR & SSIM & GMSD $\downarrow$ & HaarPSI\\
\noalign{\smallskip}
\hline
\noalign{\smallskip}
VGG           & 27.26           & 0.806   & 0.0740          & 0.7358  \\ 
L1            & 27.68           & 0.810   & 0.0753          & 0.7374  \\ 
Canny+VGG     & 27.41           & 0.810   & \textbf{0.0739} & \textbf{0.7384}  \\ 
Canny+L1      & \textbf{27.73}  & 0.810   & 0.0750          & 0.7380  \\ \hline
\end{tabular}
\end{center}
\caption{Performance of AR with different discriminators and loss functions, evaluated on the Y channel (Luminance) for LIVE1 dataset. The numbers in bold signifies the best performance. For GMSD, lower value is better.}
\label{table:fineTuning}
\end{table}
\setlength{\tabcolsep}{1.4pt}

\begin{figure}[t!]
\centering
\begin{minipage}{.15\textwidth}
  \centering
  \includegraphics[width=0.85\linewidth]{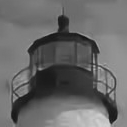}
  \small{ARCNN}
  \figfigvspace
\end{minipage}%
\begin{minipage}{.15\textwidth}
  \centering
  \includegraphics[width=0.85\linewidth]{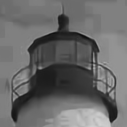}
  \small{L04}
  \figfigvspace
\end{minipage}
\begin{minipage}{.15\textwidth}
  \centering
  \includegraphics[width=0.85\linewidth]{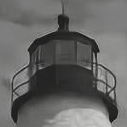}
  \small{IEGAN (B+W)}
  \figfigvspace
\end{minipage}%

\begin{minipage}{.15\textwidth}
  \centering
  \includegraphics[width=0.85\linewidth]{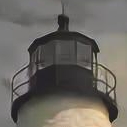}
  \small{IEGAN (RGB)}
  \figfigvspace
\end{minipage}
\begin{minipage}{.15\textwidth}
  \centering
  \includegraphics[width=0.85\linewidth]{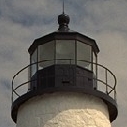}
  Ground Truth
  \figfigvspace
\end{minipage}%
\figfigvspace
\caption{Results of JPEG AR for different algorithms. The Ground Truth was degraded to 10\% of its original quality. Note that for IEGAN, the image is sharper. The IEGAN B+W (black and white) image is provided for fair comparison with the rest of the images. Best viewed in pdf.}
\label{fig:ARexample}
\end{figure}

\setlength{\tabcolsep}{4pt}
\begin{table}[]
\begin{center}
\begin{tabular}{lllll}
\hline\noalign{\smallskip}
\rowcolor[HTML]{dfdfdf} 
Algorithm & PSNR & SSIM & GMSD $\downarrow$ & HaarPSI\\
\noalign{\smallskip}
\hline
\noalign{\smallskip}
ARCNN \cite{ARCNN}   & \textbf{29.13} & 0.8232          & 0.0721          & 0.7363  \\
L4 \cite{L4L8}       & 29.08          & \textbf{0.8240} & 0.0711          & 0.7358  \\
IEGAN (Ours)   & 27.31          & 0.8124          & \textbf{0.0685} & \textbf{0.7533}  \\
\hline
\end{tabular}
\end{center}
\caption{
Performance of IEGAN for JPEG AR compared to other state-of-the-art algorithms for the LIVE1 dataset. For GMSD, lower value is better.
}
\label{table:compareAR}
\end{table}
\setlength{\tabcolsep}{1.4pt}

\begin{figure}[t!]
\fourfigures{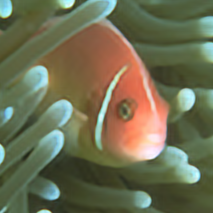}{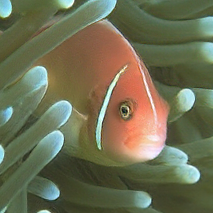}{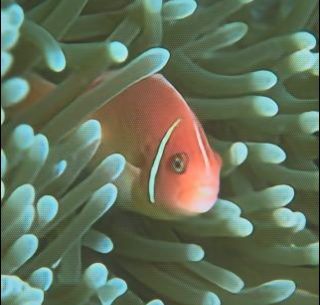}{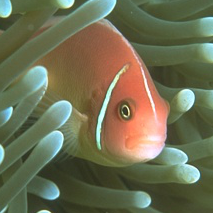}{0.21\columnwidth}
\figfigvspace
\fourfigures{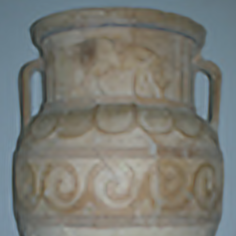}{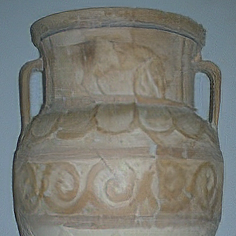}{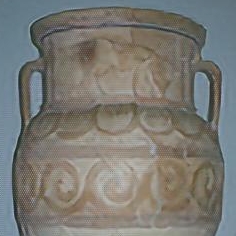}{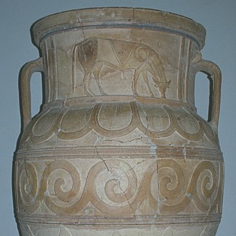}{0.21\columnwidth}
\figfigvspace
\figfigvspace
\caption{Results of SR for different algorithms. The perceptual quality of SRGAN and IEGAN outputs are visually comparable. Left to Right - SRCNN, SRGAN, IEGAN, Ground Truth. Best viewed in pdf.}
\label{fig:BSD100SR}
\end{figure}

\setlength{\tabcolsep}{4pt}
\begin{table}
\begin{center}
\begin{tabular}{lllll}
\hline\noalign{\smallskip}
\rowcolor[HTML]{dfdfdf} 
Algorithm & PSNR & SSIM & GMSD $\downarrow$ & HaarPSI\\
\noalign{\smallskip}
\hline
\noalign{\smallskip}
SRCNN \cite{SRCNN}& \textbf{27.04} & \textbf{0.784} & 0.088 & 0.713  \\
SRGAN \cite{SRGAN}& 26.02 & 0.740 & 0.086 & 0.728 \\
ENet-PAT \cite{enhancenet} & 25.77  & 0.718  & 0.088  & 0.719 \\
IEGAN (Ours)& 25.03  & 0.735  & \textbf{0.085} & \textbf{0.730}\\
\hline
\end{tabular}
\end{center}
\caption{
Performance of state-of-the-art algorithms for SR for Set14 dataset for RGB images. For GMSD lower, value is better.
}
\label{table:compareSR}
\end{table}
\setlength{\tabcolsep}{1.4pt}

\setlength{\tabcolsep}{4pt}
\begin{table}[]
\begin{center}

\begin{tabular}{llllll}
\hline\noalign{\smallskip}
& & \textbf{LIVE1} & &  \\
\hline\noalign{\smallskip}
\rowcolor[HTML]{dfdfdf} 
Algorithm & PSNR & SSIM & GMSD $\downarrow$ & HaarPSI\\
\noalign{\smallskip}
\hline
\noalign{\smallskip}
\small{ARCNN+SRGAN}  & 21.61            & 0.5284            & 0.1980            & 0.4112  \\
\small{SRGAN+ARCNN}  & \textbf{22.70}   & \textbf{0.6417}   & 0.1457            & 0.5302  \\
\small{IEGAN}      & 22.57            & 0.6319            & \textbf{0.1404}   & \textbf{0.5504}  \\

\hline\noalign{\smallskip}
& & \textbf{World} & \textbf{100} &  \\
\hline\noalign{\smallskip}
\rowcolor[HTML]{dfdfdf} 
Algorithm & PSNR & SSIM & GMSD $\downarrow$ & HaarPSI\\
\noalign{\smallskip}
\hline
\noalign{\smallskip}

\small{ARCNN+SRGAN}   & 25.51            & 0.6809             & 0.1668             & 0.4792  \\
\small{SRGAN+ARCNN}   & \textbf{27.16}   & \textbf{0.7861}    & 0.1059             & 0.6320  \\
\small{IEGAN}      & 25.62            & 0.7651             & \textbf{0.1009}    & \textbf{0.6429}  \\

\hline
\end{tabular}
\end{center}
\caption{
Performance of IEGAN for simultaneous AR+SR compared to other state of the art algorithms for the benchmark LIVE1 dataset and the World100 dataset. For GMSD, lower value is better.
}
\label{table:compareARSR}
\end{table}
\setlength{\tabcolsep}{1.4pt}

\begin{figure*}[]
\begin{center}

\begin{minipage}{.21\textwidth}
  \centering
  \includegraphics[width=0.95\linewidth]{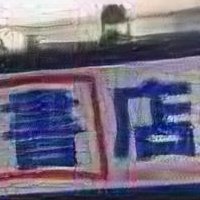}
  \figfigvspace

\end{minipage}%
\begin{minipage}{.21\textwidth}
  \centering
  \includegraphics[width=0.95\linewidth]{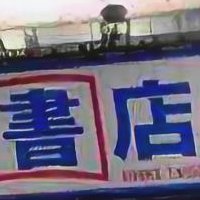}
  \figfigvspace

\end{minipage}%
\begin{minipage}{.21\textwidth}
  \centering
  \includegraphics[width=0.95\linewidth]{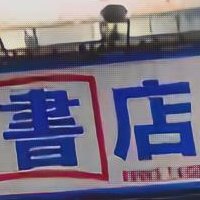}
  \figfigvspace

\end{minipage}%
\begin{minipage}{.21\textwidth}
  \centering
  \includegraphics[width=0.95\linewidth]{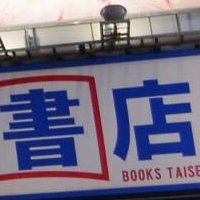}
  \figfigvspace

\end{minipage}%

\begin{minipage}{.21\textwidth}
  \centering
  \includegraphics[width=0.95\linewidth]{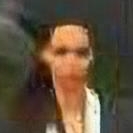}
  \figfigvspace

\end{minipage}%
\begin{minipage}{.21\textwidth}
  \centering
  \includegraphics[width=0.95\linewidth]{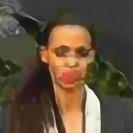}
  \figfigvspace

\end{minipage}%
\begin{minipage}{.21\textwidth}
  \centering
  \includegraphics[width=0.95\linewidth]{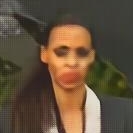}
  \figfigvspace

\end{minipage}%
\begin{minipage}{.21\textwidth}
  \centering
  \includegraphics[width=0.95\linewidth]{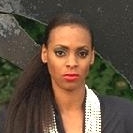}
  \figfigvspace

\end{minipage}%

\begin{minipage}{.21\textwidth}
  \centering
  \includegraphics[width=0.95\linewidth]{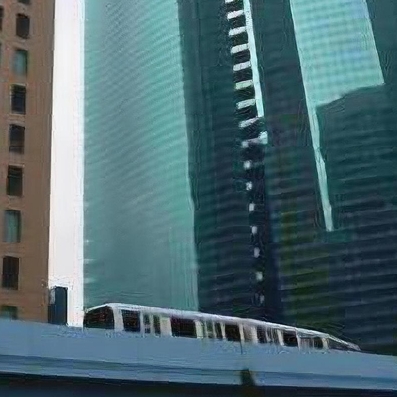}
  \figfigvspace

\end{minipage}%
\begin{minipage}{.21\textwidth}
  \centering
  \includegraphics[width=0.95\linewidth]{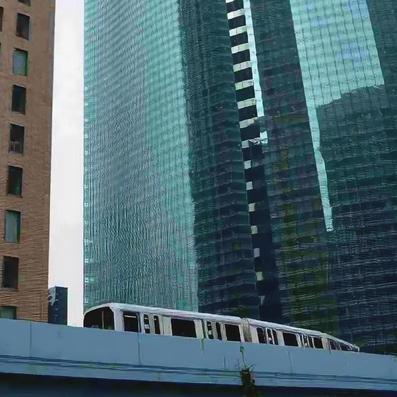}
  \figfigvspace

\end{minipage}%
\begin{minipage}{.21\textwidth}
  \centering
  \includegraphics[width=0.95\linewidth]{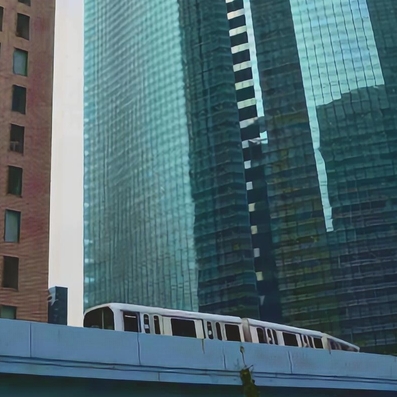}
  \figfigvspace

\end{minipage}%
\begin{minipage}{.21\textwidth}
  \centering
  \includegraphics[width=0.95\linewidth]{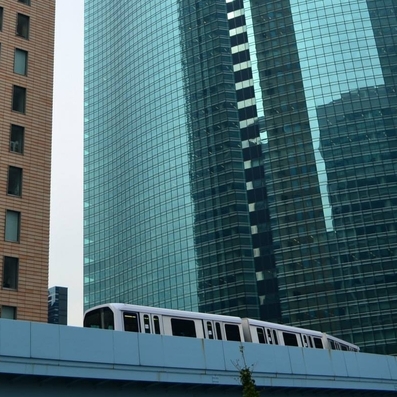}
  \figfigvspace

\end{minipage}%

\begin{minipage}{.21\textwidth}
  \centering
  \includegraphics[width=0.95\linewidth]{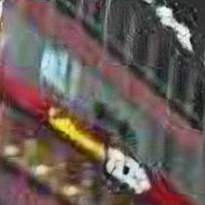}
  \figfigvspace
  
  SRGAN+ARCNN
\end{minipage}%
\begin{minipage}{.21\textwidth}
  \centering
  \includegraphics[width=0.95\linewidth]{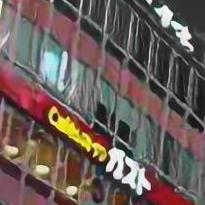}
  \figfigvspace
  
  ARCNN+SRGAN
\end{minipage}%
\begin{minipage}{.21\textwidth}
  \centering
  \includegraphics[width=0.95\linewidth]{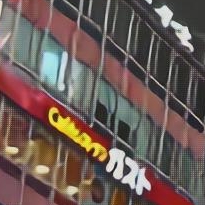}
  \figfigvspace
  
  \textbf{IEGAN}
\end{minipage}%
\begin{minipage}{.21\textwidth}
  \centering
  \includegraphics[width=0.95\linewidth]{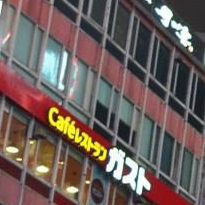}
  \figfigvspace

  Ground Truth
\end{minipage}%

\figfigvspace
\end{center}
\caption{Results for simultaneous SR+AR of RGB images using various algorithms. Row 1, 3 \& 4 are from World100 and row 2 from LIVE dataset. Note that the textures and details in the output images from IEGAN are much superior than the others. Best viewed in pdf.}
\label{fig:ARSR}
\end{figure*}

\subsection{Ablation Study}
\label{subsec:investigation}
We investigate the effect of different discriminator architectures and loss functions on the performance of our network. 

\textbf{Discriminator: }
We use two different discriminators in our experiments. The first discriminator (Dv1) evaluates the image in the pixel space as described in Section \ref{sec:method} and the other one (Dv2) in the feature space which gives a binary output of 0 or 1 for fake and real images respectively. Architecture of Dv1 is shown in Figure \ref{fig:architecture}. For Dv2, we use a similar discriminator introduced by Ledig {\it et al.} \cite{SRGAN} in SRGAN. We train all the model by converting all the images to YCbCr color space. Since the human visual system has poor frequency response to color components (CbCr) compared to luminance (Y), we try to minimize most of the artifacts in the Y channel for best perceptual quality.

\textbf{Loss Function: } We use a weighted combination of the VGG feature maps with the Canny edge detector as loss function. We also study the performance using VGG and L1 separately combined with Canny to validate our claim that combining Canny enhances the perceptual quality of the images. Table \ref{table:fineTuning} shows the quantitative performance of the algorithm with various discriminator and loss functions. We also experiment with Holistically-Nested Edge Detection (HED) \cite{hed} by replacing the Canny counterpart. HED is the state of the art edge detection algorithm which has better edge detection capabilities compared to Canny. 
However, from Figure \ref{fig:edge}, we can observe that both HED and Canny successfully produce the required edge information, which is perceptually indistinguishable to the human eye.
In other words, the different edge methods are not critical to the overall performance, which is also proved in Table \ref{table:combiResults}.
Thus we choose the simple yet fast Canny method in our final framework.

The results from Table \ref{table:combiResults} and Table \ref{table:fineTuning} confirm that the GAN with discriminator Dv1 using a weighted combination of VGG with the Canny loss function gives the best GMSD and HaarPSI score. Majority of the AR algorithms proposed till date works on the black and white images. Our proposed algorithm works with color images as well. The highest PSNR and SSIM values are obtained from the framework having the discriminator Dv1 with the Canny+L1 loss function. PSNR is a common measure used to evaluate AR and SR algorithms. However, the ability of PSNR to capture perceptually relevant differences is very limited as they are defined based on pixel-wise image differences \cite{SRGAN,structuralSimIQA,structuralSimIQA2}. We will be using Dv1 with loss VGG + Canny for the rest of the experiments. We name this framework as IEGAN.

\subsection{Comparison to State of the Art}
\label{subsec:final_algo}
From Table \ref{table:compareAR}, we see that for JPEG artifact removal purpose, IEGAN performs significantly better than other algorithms. According to the results of Table \ref{table:compareSR}, IEGAN gives the best GMSD and HaarPSI scores for Set14.
For SR, IEGAN produces comparable results to SRGAN but the perceptual quality of the images generated by SRCNN is much inferior. This has been demonstrated in Figure \ref{fig:BSD100SR} with two images from the BSD100 dataset. Furthermore, Ledig {\it et al.} \cite{SRGAN} has also argued that the perceptual quality of the images generated by SRCNN is not good comparing to SRGAN by mean opinion score (MOS) testing. 

The results for end-to-end AR+SR are shown in Table \ref{table:compareARSR} where we can see that IEGAN outperforms the other state of the art pipelines. Figure \ref{fig:ARSR} shows the visual result on parts of the \textbf{World100} dataset where algorithms were used to perform both AR and SR on the same image simultaneously. ARCNN+SRGAN implies that first ARCNN was used to recover the image from artifacts and then SRGAN was used to super-resolve the image, while SRGAN+ARCNN implies that first SRGAN was used, and then ARCNN. In contrast, IEGAN provides a one-shot end-to-end solution for both AR and SR in the same network. However, all the algorithms fail to produce the photo-realistic result for large areas having a very gentle color gradient, e.g., clear sky, aurora etc. The images in the World100 dataset are more than 2000 pixels on at least one side. Thus the area of the color gradient becomes enormously large compared to the receptive field of the network. The network is trained with 128$\times$128 images and the training data hardly contains any image with such color gradient, resulting in a lack of training for such images. Thus the algorithm fails to learn how to recreate a smooth color gradient in the output images. This fact is also true for all the other algorithms too. 

\section{Conclusion}
We have described a deep generative adversarial network with skip connections that sets a new state of the art on public benchmark datasets when evaluated with respect to perceptual quality. This network is the first framework which successfully recovers images from artifacts and at the same time super-resolves, thus having a single-shot operation performing two different tasks. We have highlighted some limitations of the existing loss functions used for training any image enhancement network and introduced IEGAN, which augments the feature loss function with an edge loss during training of the GAN. Using different combinations of loss functions and by using the discriminator both in feature and pixel space, we confirm that IEGAN reconstructions for corrupted images are superior by a considerable margin and more photo-realistic than reconstructions obtained by the current state-of-the-art methods.

{\small
\bibliographystyle{ieee}
\bibliography{egbib}
}

\end{document}

%% file: WACV_IEGAN/macro.tex
\usepackage{color}
\usepackage[normalem]{ulem}

\def\figfigvspace{\vspace{0.18cm}}

\newcommand{\fourfigures}[5]{
            \centerline{
              $\vcenter{\hbox{{\includegraphics[width=#5]{#1}}}}$~~
              $\vcenter{\hbox{{\includegraphics[width=#5]{#2}}}}$~~
              $\vcenter{\hbox{{\includegraphics[width=#5]{#3}}}}$~~
              $\vcenter{\hbox{{\includegraphics[width=#5]{#4}}}}$}}